\documentclass[10pt,twocolumn,letterpaper]{article}

\usepackage[pagenumbers]{cvpr} 

\usepackage{caption}
\usepackage{subcaption}
\usepackage{graphicx}
\usepackage{amsmath}
\usepackage{amssymb}
\usepackage{booktabs}
\usepackage{breqn}
\usepackage{tablefootnote}
\usepackage{multirow}
\usepackage{makecell}
\usepackage{tablefootnote}
\usepackage{footnote}
\usepackage[normalem]{ulem}

\usepackage[pagebackref,breaklinks,colorlinks]{hyperref}

\usepackage[dvipsnames]{xcolor}
\definecolor{aquamarine}{HTML}{00B5BE}

\newcommand{\name}{{Swin-Free}\xspace}

\def\cvprPaperID{00} 
\def\confName{CVPR}
\def\confYear{2022}

\begin{document}

\title{\name: Achieving Better Cross-Window Attention and Efficiency with Size-varying Window}

\author{Jinkyu Koo, John Yang, Le An, Gwenaelle Cunha Sergio, and Su Inn Park\\
NVIDIA, 2788 San Tomas Expy, Santa Clara, CA 95051\\
{\tt\small \{jinkyuk, johnyang, lean, gcunhasergio, joshp\}@nvidia.com}
}
\maketitle

\begin{abstract}
Transformer models have shown great potential in computer vision, following their success in language tasks. Swin Transformer is one of them that outperforms convolution-based architectures in terms of accuracy, while improving efficiency when compared to Vision Transformer (ViT) and its variants, which have quadratic complexity with respect to the input size. 
Swin Transformer features shifting windows that allows cross-window connection while limiting self-attention computation to non-overlapping local windows. 
However, shifting windows introduces memory copy operations, which account for a significant portion of its runtime.
To mitigate this issue, we propose \name in which we apply size-varying windows across stages, instead of shifting windows, to achieve cross-connection among local windows.
With this simple design change, \name runs faster than the Swin Transformer at inference with better accuracy. Furthermore, we also propose a few of \name variants that are faster than their Swin Transformer counterparts.
\end{abstract}

\begin{figure*}[t]
\centering
\hspace*{\fill}
\begin{subfigure}{\textwidth}
    \centering
    \captionsetup{justification=centering}
    \includegraphics[width=\textwidth]{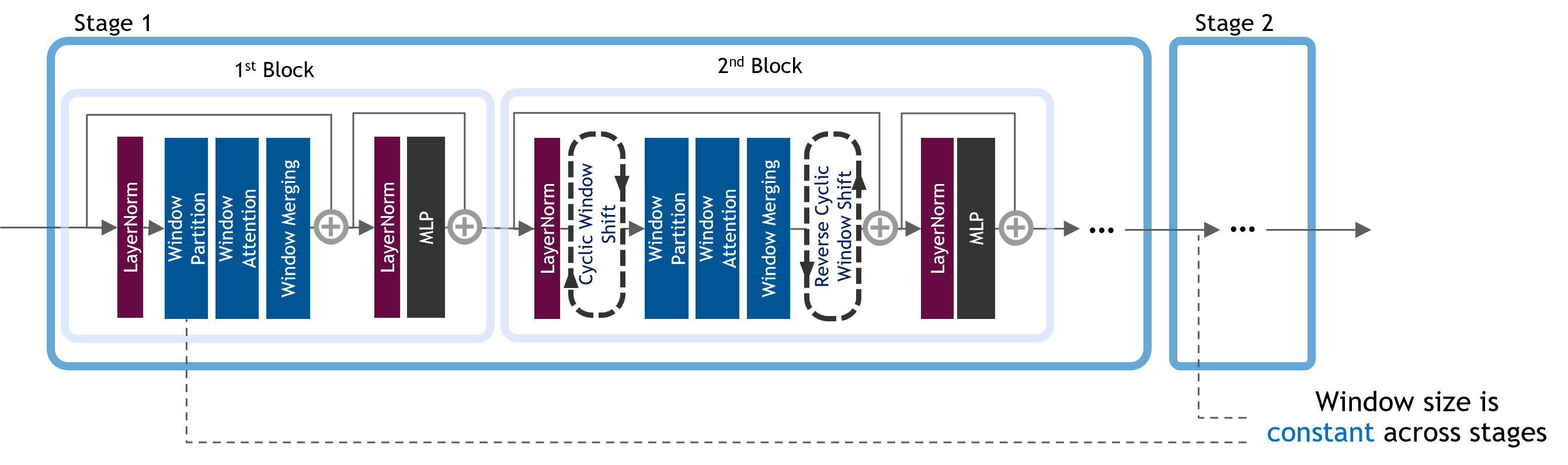}
    \caption{Conventional window-shifting block structures of Swin transformers \cite{liu2021swin}}
    \label{fig:swin_block}
\end{subfigure} \\
\hfill
\begin{subfigure}{\textwidth}
    \centering
    \captionsetup{justification=centering}
    \includegraphics[width=\textwidth]{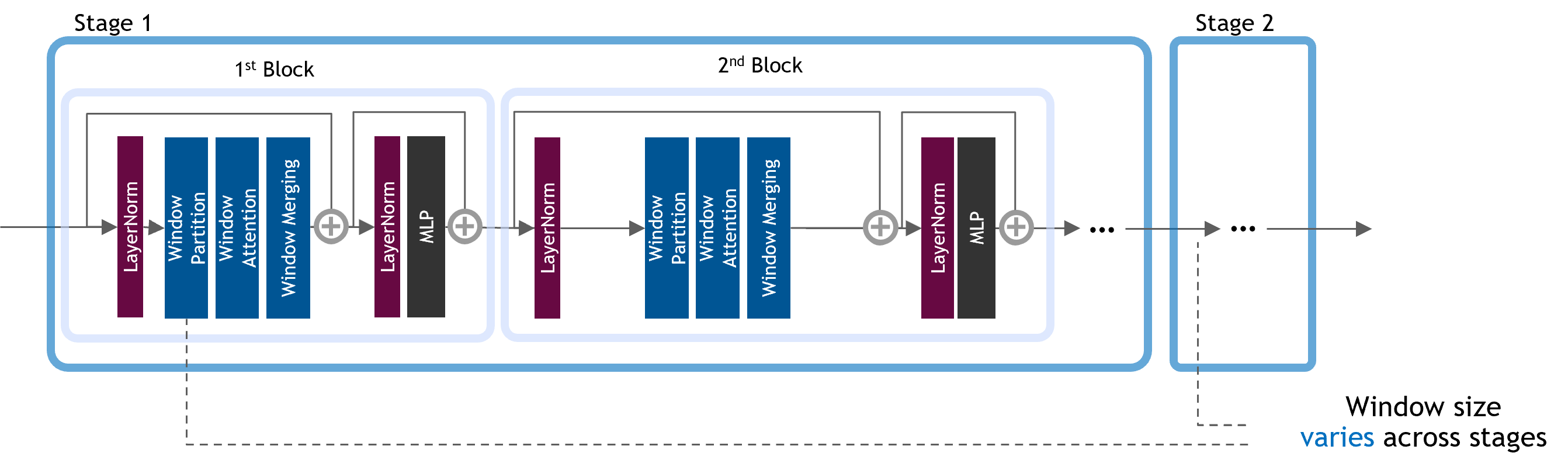}
    \caption{Our proposed block structures with varying window sizes}
    \label{fig:swinnext_block}
\end{subfigure}
\caption{Comparison in functional blocks between Swin and \name. Note that in \name, shifting windows is removed and the size of the local window varies across stages.}
\label{fig:comp_arch}
\end{figure*}

\section{Introduction}\label{sec:intro}

Until recently, convolutional neural network (CNN) had been leading the remarkable innovations in computer vision tasks which had otherwise been considered too difficult in the past, such as autonomous driving \cite{NIPS2012_c399862d,DBLP:journals/corr/SimonyanZ14a,DBLP:journals/corr/SzegedyLJSRAEVR14,DBLP:journals/corr/HeZRS15,DBLP:journals/corr/HanMD15,DBLP:journals/corr/HowardZCKWWAA17}. However, the leading role of CNNs is recently being transferred to Transformer-based networks~\cite{vit,liu2021swin,DBLP:journals/corr/abs-2111-09883}. The Transformer model was first proposed for natural language processing (NLP) tasks, such as text classification and machine translation, and it has demonstrated great success \cite{DBLP:journals/corr/abs-1810-04805,radford2018improving,DBLP:journals/corr/abs-2005-14165}. Such a breakthrough in the language domain has sparked great interest in the computer vision community and recently lead to promising results on various tasks such as image classification~\cite{vit, liu2021swin} and semantic segmentation~\cite{xie2021segformer}.

The key component in Transformer architecture is the self-attention module, which learns the relevance of one element to the other elements of a sequence.
Unlike recurrent networks, such as LSTM~\cite{HochSchm97}, that can only attend to context within a limited scope, the self-attention mechanism
explicitly models the interactions among all entities of a sequence. This allows Transformers to learn global context at once, resulting in their success in many applications~\cite{DBLP:journals/corr/abs-2005-14165,chowdhery2022palm,vit}. A drawback is, however, that computation complexity of the self-attention increases quadratically with respect to the length of an input sequence. This can be a critical problem especially in computer vision tasks, since the sequence length, often determined by the image resolution, can be intractably large.

Swin Transformer~\cite{liu2021swin} mitigates the quadratic complexity issue by partitioning an image into non-overlapping windows and computing self-attention within the local windows.
To bridge the non-overlapping windows, Swin Transformer features shifting the window partition between consecutive self-attention layers, providing cross-connections among local windows. While this design choice leads to improved efficiency and accuracy, the operations for shifting windows incur data movement in memory. In fact, as shown in Table~\ref{table:op_profile}, shifting windows account for about 8.7$\%$ of the total runtime for a Swin Transformer model, when inference is performed with NVIDIA TensorRT \cite{tensorrt}.

To mitigate this shortcoming of Swin Transformer, we propose \name, which does not shift local windows in order to reduce data movement. Instead, to achieve cross-connection among non-overlapping windows, \name varies the size of windows across different stages (see Table \ref{table:comp_arch}). For example, \name may double the window size at a stage in order to model cross-attention among smaller local windows of the previous stage.

\begin{table}[t]
\caption{Operation profile of a Swin Transformer model (Swin-B) on NVIDIA RTX 3080 GPU.}\label{table:op_profile}
\centering
\begin{tabular}{ccc}
\hline
\multicolumn{1}{c}{\multirow{2}{*}{Operation}} & \multicolumn{2}{c}{Percentage ($\%$) in runtime}                             \\ \cline{2-3} 
\multicolumn{1}{c}{}                                & \makecell{TensorRT\\(FP16)} & \makecell{PyTorch\\(FP32)} \\ \midrule\midrule[.1em]
Shifting windows    &  8.74  &  4.39       \\ 
LayerNorm           &  10.11 &  9.63       \\ 
GELU                &  13.46 &  3.15       \\ \hline
\end{tabular}
\end{table}

Experimental results show that \name featuring the size-varying windows reduces the model runtime significantly as compared to Swin Transformer, mainly thanks to avoiding shifting windows and being able to leverage faster matrix multiplication with larger inputs.
Note that on modern GPUs, efficient implementations of math operations such as convolution with large kernels are widely available. In \name, a larger portion of its runtime is spent on computation rather than memory copy, indicating a better GPU utilization.
At the same time, \name improves the classification accuracy as well, implying that the size-varying windows can provide better modeling power than shifting windows with a constant window size.

We also propose several variants of \name that prioritize latency over accuracy. In other words, with on par accuracy, a variant of \name is designed to be faster than its Swin Transformer counterpart. In addition, we further simplify \name with more efficient layers such as BatchNorm and ReLU, instead of more commonly used but expensive LayerNorm and GELU layers, which also account for significant part of the runtime (see Table \ref{table:op_profile}). With those design elements, we were able to improve the latency by 19\% compared to Swin-B. In addition, we also show that by utilizing the improved modeling power of \name, we can further reduce the depth of our model. For example, a variant of \name is faster than Swin by about 33\% without loss of accuracy (see Table \ref{table:exp1k}).

\section{Related Work}\label{sec:related}
\textbf{Convolutional Neural Network (CNN):}
Over the past decade, CNNs have been the \textit{de facto} standard in computer vision, and keep improving accuracy with innovations in architecture design \cite{NIPS2012_c399862d,DBLP:journals/corr/SimonyanZ14a,DBLP:journals/corr/SzegedyLJSRAEVR14,DBLP:journals/corr/HeZRS15}.
In parallel, a lot of efforts have also been made to reduce the complexity of CNN models for efficiency. Such directions include model compression, quantization, and low cost operations such as depth-wise convolution~\cite{DBLP:journals/corr/HanMD15,DBLP:journals/corr/HowardZCKWWAA17}.
Although CNNs are still dominant in computer vision tasks, many recent works have demonstrated that Transformer-based models outperform the state-of-the-art CNN-based models \cite{vit,liu2021swin,DBLP:journals/corr/abs-2111-09883}. Arguably, we are about to see a paradigm shift in computer vision from CNN to Transformer.

\textbf{Transformer Architectures:}
Introduced in a pioneer work \cite{vaswani2017attentionisallyouneed} for machine translation tasks, Transformers have become the state-of-the-art models for NLP tasks, replacing most of the LSTM-based sequence-to-sequence approaches \cite{DBLP:journals/corr/abs-1810-04805, DBLP:journals/corr/abs-1910-10683,DBLP:journals/corr/abs-1901-02860,radford2018improving,DBLP:journals/corr/abs-2005-14165}.
As opposed to recurrent networks that process short-term context recursively, Transformer architectures are based on the attention mechanism, which explicitly models the relative importance among all elements of a sequence, thereby learning sequence-wide relationships. In other words, Transformers process a sequence as a whole and recursion is totally avoided.

\textbf{Transformer in vision:}
With minimal vision-specific modifications, ViT \cite{vit} applies the attention mechanism to image classification tasks.
As the counterpart of input token embeddings, ViT divides the images into patch embedding sequences and feeds them into a standard Transformer.
ViT outperforms CNNs in image classifications, but it has been often reported to be difficult to train compared to CNNs.
Since the computational complexity of the attention operation is quadratically proportional to the input size, ViT has challenges to take high-resolution images in as inputs.
Other Transformer-based vision models such as DETR \cite{DETR} and SETR \cite{SETR} also hold such a quadratic complexity issue.

\section{Preliminary: Swin Transformer}

\begin{figure}[t]
    \centering
    \begin{subfigure}{0.4\linewidth}
        \centering
        \includegraphics[width=0.8\textwidth]{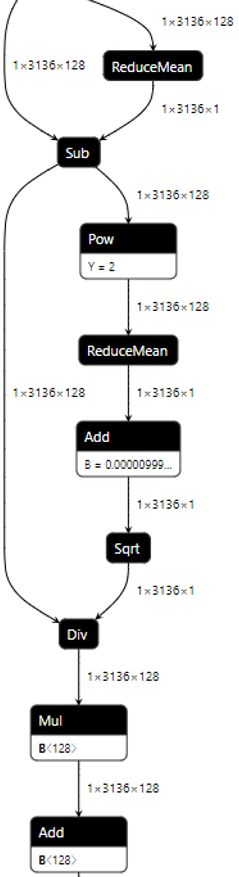}        
        \caption{LayerNorm}
        \label{fig:layernorm}
    \end{subfigure}
    \begin{subfigure}{0.4\linewidth}
        \centering
        \includegraphics[width=0.9\textwidth]{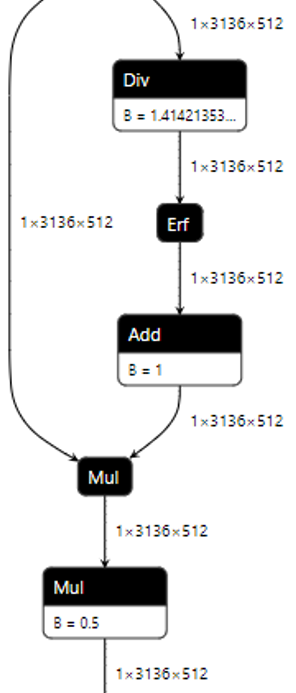}
        \caption{GELU}
        \label{fig:gelu}
     \end{subfigure}
    \caption{Examples of ONNX representations of LayerNorm and GELU.}
    \label{fig:layernorm_gelu}
\end{figure}

Swin Transformer \cite{liu2021swin} leverages a multi-stage hierarchical architecture, where the input image is first divided into small-sized patches and feature maps are gradually merged with neighboring patches along the stages.
With these hierarchical representations, Swin Transformer can easily be applied to dense prediction tasks such as object detection and segmentation.
Swin Transformer achieves a linear computational complexity by computing self-attention within non-overlapping local windows.
To capture interactions between local windows, the shifted window scheme that alternates between two window configurations in consecutive Transformer blocks is employed.

Shifting windows plays a critical role in achieving Swin Transformer's claimed accuracy, but also introduces a lot of memory movements. As shown in Table \ref{table:op_profile}, the shifting window operations in Swin-B (one of Swin Transformer variants) account for 8.7\% of the total runtime with NVIDIA TensorRT (FP16 precision) and 4.4\% with PyTorch (FP32 precision). This suggests that there is room for latency improvement if memory movements can be minimized.

In addition, LayerNorm and GELU used in Swin Tranformer are also responsible for a significant portion of the runtime as shown in Table \ref{table:op_profile}. Taking a look at those two operations in ONNX representation~\cite{bai2019} in Figure \ref{fig:layernorm_gelu}, a cascade of math operations can be identified to fulfill those two layers. Previous study has suggested that by strategically using BatchNorm and ReLU layers, the accuracy of a Transformer model will not be degraded much~\cite{dest}. In this paper, we attempt to improve on top of Swin Transformer for both accuracy and runtime, and propose \name, which will be explained in the following section.

\section{Method}\label{sec:method}

\subsection{Overview of \name}\label{sec:overall_arch}
\begin{figure*}[t]
    \centering
    \includegraphics[width=0.95\linewidth]{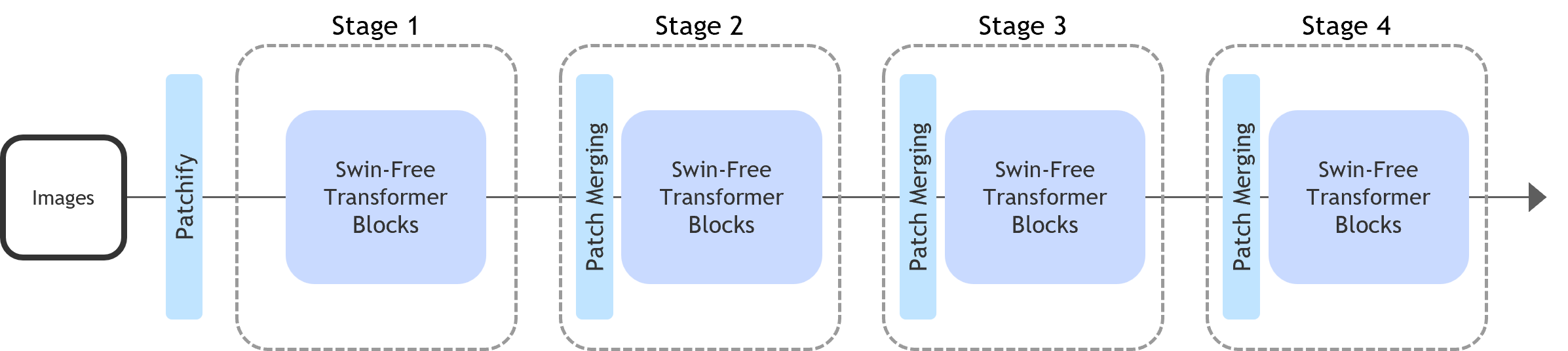}
    \caption{Overall architecture of \name.}
    \label{fig:arch}
\end{figure*}
\begin{table}[t]
\caption{Comparison between Swin and \name for the input size 224$\times$224. Here, $P$ means the number of patches at the beginning of a stage. The values of $M$ and $N$ denote the size of a local window and the number of non-overlapping windows in a stage, respectively. Note that Swin Transformer applies shifting windows in every other Transformer block, while \name does not shift windows.}\label{table:comp_arch}
\centering
\begin{tabular}{ccc}
\hline
Stage & Swin                       & \name \\ \midrule\midrule[.1em]
1     & \makecell{$P=56\times56$\\ $M=7$\\ $N=64$} &  \makecell{$P=56\times56$\\ $M=7$\\ $N=64$}   \\ \hline
2     & \makecell{$P=28\times28$\\ $M=7$\\ $N=16$} &  \makecell{$P=28\times28$\\ $M=14$\\ $N=4$}   \\ \hline
3     & \makecell{$P=14\times14$\\ $M=7$\\ $N=4$} &  \makecell{$P=14\times14$\\ $M=14$\\ $N=1$}   \\ \hline
4     & \makecell{$P=7\times7$\\ $M=7$\\ $N=1$} &  \makecell{$P=7\times7$\\ $M=7$\\ $N=1$}   \\ \hline
\end{tabular}
\end{table}

Our baseline architecture shown in Figure \ref{fig:arch} is similar to Swin Transformer \cite{liu2021swin}, except that it does not use the shifted windows. The input image is first patchified. Each stage applies a number of Swin-style Transformer blocks for the patches, where the self-attention computation is done within each of non-overlapping local windows. Here, the local window operates on an $M\times M$ patch. Like in Swin Transformer, the number of patches are reduced by half at each stage by the patch merging layer. The only difference from Swin Transformer is that we do not shift the local windows. Instead, we choose to  vary the size of the local window (\textit{i.e.}, $M$) at each stage, which will be explained in more detail in Section \ref{sec:size_varying_windows}. 

The difference between Swin and \name for input size $224\times224$ is summarized in Table \ref {table:comp_arch}. Note that in stage 2 and 3, \name uses a larger window size than Swin, and therefore the number of non-overlapping windows in \name is smaller at those stages than in Swin. Figure \ref{fig:comp_arch} also shows how \name is different from Swin in detail at the block level. In \name, shifting windows and its reverse operation used in Swin Transformer are removed, and the size of the window changes with each stage.

\subsection{Size-Varying Windows}\label{sec:size_varying_windows}

Shifting the local windows in Swin Transformer is an effective way to achieve cross-connection among windows, but it requires moving data in memory. This is typically more costly than math computation on GPUs, and can therefore negatively impact the model efficiency. In fact, as shown in Table \ref{table:op_profile}, shifting windows takes a considerable portion of the total runtime.

To avoid using the shifted windows, we enable cross-connection between non-overlapping windows by changing the size of the local windows at each stage. Recall that $M$ is the size of the local window. As Table \ref{table:comp_arch} shows, in our implementations for the input size 224$\times$224, we vary the value of $M$ as $M=7,14,14,7$ for the four stages.
From this setup, we consider the cross-connection among four neighboring 7$\times$7 local windows at stages 2 and 3, \textit{i.e.}, a 14$\times$14 local window in the current stage effectively includes four of 7$\times$7 local windows in the previous stage. 

The above changes may increase GPU computation load of a single local window due to the enlarged window size in the attention block. However, note in Table \ref{table:comp_arch} that the number of non-overlapping local windows (\textit{i.e.}, $N$) in stages 2 and 3 of \name becomes one fourth of that in Swin. In other words, in the matrix multiplication of \name, the matrices' size is larger, but the number of matrices to be processed is smaller.
We have observed that processing a 14$\times$14 local window does not increase the latency as compared to processing four of 7$\times$7 local windows on GPU, but rather decreased the latency, thanks to their massive parallel computing capability. We will discuss this point in more detail in Section \ref{sec:experiments}.

\subsection{Further Optimization}\label{sec:variants}

\textbf{Replacement of LayerNorm and GELU:}
As shown in Figure \ref{fig:layernorm_gelu}, LayerNorm and GELU are composed of multiple math layers, which require more computation as compared to the commonly used BatchNorm and ReLU layers. In Table \ref{table:op_profile}, it is observed that LayerNorm and GELU account for about 24\% of the total runtime of a Swin Transformer model when running with TensorRT. Thus, when the latency is also critical in an application, we replace them with BatchNorm and ReLU without significant accuracy degradation\cite{dest}. It can be seen in Section \ref{sec:experiments} that such modification allows \name to run even faster while still surpassing Swin Transformer in terms of accuracy.

\textbf{Depth reduction:}
Another way to prioritize latency is to reduce the depth of a model. Specifically, we consider reducing the number of Transformer blocks at stage 3. For example, compared to Swin-B, where stage 3 consists of 18 Transformer blocks, we may consider using 14 blocks only. We will see in Section \ref{sec:experiments} that this variant of \name can still achieve better accuracy than Swin Transformer with significant improvement in latency.

\section{Experiments}\label{sec:experiments}

\begin{table*}[t]
\caption{Model variants: (a) We consider variants by changing hyper-parameters of a given architecture. (b) We apply architectural modification to a given model. The abbreviated symbol of each variant is added to a model name as a postfix.}\label{table:variant1}\label{table:variant}
\centering
\begin{subtable}{0.90\linewidth}
\caption{Variants by hyper-parameters.}\label{table:variant1}
\centering
\begin{tabular}{ccc}
\hline
Variant & \makecell{Embedding dimension per patch} & \makecell{\# of blocks at a stage (depth)} \\ \midrule\midrule[.1em]
Tiny (T)  & 96  & \{2,2,6,2\}     \\ 
Small (S) & 96  & \{2,2,18,2\} \\ 
Base (B) & 128  & \{2,2,18,2\} \\ \hline
\end{tabular}
\end{subtable}
\vspace{3mm}

\begin{subtable}{0.90\linewidth}
\caption{Variants by modification.}\label{table:variant2}
\centering
\begin{tabular}{cc}
\hline
Variant & Modification \\ \midrule\midrule[.1em]
BatchNorm/ReLU (BR)	&	Replace LayerNorm with BatchNorm and GELU with ReLU. \\
Depth reduction to $x$ (DR$x$)	&	Reduce the number of Transformer blocks at stage 3 to $x$. \\ \hline
\end{tabular}
\end{subtable}
\end{table*}

Our focus of experiments is to compare \name with Swin Transformer in terms of both latency and accuracy in classification tasks. All latency results are measured using NVIDIA RTX 3080, PyTorch 1.13, and TensorRT 8.5.3 with CUDA 11.8. Evaluations are done with the ImageNet dataset \cite{deng2009imagenet} with 1K classes and input shape 224$\times$224. We consider the same variant models as in Swin Transformer, shown in Table \ref{table:variant1}. Note that we do not consider the Large (L) variant with embedding dimension 192 used in Swin, since it requires what is called the fall11 version of the 22K-class dataset that is no longer available.
Like Swin-B, we add a post-fix to a model name to indicate its variant (\textit{e.g.}, \name-B). 
Additionally, we also consider other variants resulting from the modification in Table \ref{table:variant2}, mentioned in Section \ref{sec:variants}. These additional optimizations enhance the latency of a model, possibly at the cost of reduced accuracy. The abbreviated symbols of these variants  (\textit{i.e.}, BR or DR$x$) are also added as a post-fix to a model name.

\subsection{Shifted windows of Swin}\label{sec:shift_on_off}

\begin{table}[]
\caption{Turning on/off shifting windows in Swin-B at each stage: 1 means `on'. For example, `1, 1, 1, 1' implies that all stages use the shifted windows, meaning exactly Swin-B. The symbol `-' means that training could not finish successfully (\textit{i.e.}, diverged). }\label{table:shifton}
\centering
\begin{tabular}{ccc}
\hline
Case & On/off on cyclic shift & Top-1 accuracy (\%) \\ \midrule\midrule[.1em]
1 & 1, 1, 1, 1	& 83.4\\ 
2 & 0, 1, 1, 1	& 82.3 \\ 
3 & 0, 0, 1, 1	& 82.3 \\ 
4 & 0, 0, 0, 1	& - \\ 
5 & 0, 0, 1, 0	& 82.2 \\ 
6 & 0, 1, 0, 0	& - \\ 
7 & 1, 0, 0, 0	& - \\ 
8 & 0, 0, 0, 0	& - \\ \hline
\end{tabular}
\end{table}

Before going into the evaluation of \name, we first want to understand the importance of the shifted window in each stage of Swin-B. Table \ref{table:shifton} shows the top-1 accuracy of Swin-B depending on which stage has shifting windows enabled or disabled. Note that Case 1 uses the shifted windows for all stages, and thus it is exactly the same as Swin-B.\footnote{Even though we used the same training configuration, our Swin-B's top-1 accuracy, trained from scratch, is 83.4\%, which is slightly lower than 83.5\% reported in \cite{liu2021swin}.} We can first see from Case 8 that without the shifted windows, it is even difficult to successfully complete training, and thus the shifted windows is indeed critical in Swin.
We can also see from Cases 4 to 7 that stage 3 is a critical stage to use the shifted windows. This is, to some extent, not surprising, since stage 3 is a dominant portion of Swin-B.
However, we can also see from Cases 1 to 3 that selectively using the shifted windows over each stage marginally helps in increasing accuracy. Thus, it is important to apply them to all stages of Swin-B.

\subsection{Windows size of \name}\label{sec:swin-next-window-size}

\begin{table*}[]
\caption{Latency and accuracy according to the variation in window size at each stage of Swin-B without using cyclic shift. For example, `7, 7, 14, 7' means that stage 3 uses 14 as the window size, while stages 1, 2, and 4 use 7. The symbol `-' means that training could not finish successfully (\textit{i.e.}, diverged).}\label{table:win}
\centering
\begin{tabular}{cccc}
\hline
Case & Window size at a stage & Top-1 accuracy (\%) & Latency in PyTorch (FP32) (ms) \\ \midrule\midrule[.1em]
1 & 7, 7, 7, 7  & - & 13.7 \\ \hline

2 & 7, 7, 14, 7 & \textbf{83.8} & 12.7 \\ 
3 & 7, 14, 7, 7 & 81.1 & 13.7 \\ 
4 & 14, 7, 7, 7 & 81.2 & 13.7 \\ \hline

5 & 7, 14, 14, 7 & \textbf{83.8} & \textbf{12.6} \\ 
6 & 14, 7, 14, 7 & \textbf{83.8} & \textbf{12.6} \\ 
7 & 14, 14, 7, 7 & 81.2 & 13.8 \\ \hline

8 & 14, 14, 14, 7 & 83.7 & \textbf{12.6} \\ \hline
\end{tabular}
\end{table*}

In this section, we show which window size configurations are better suited for each stage of \name. To ensure fair comparison with Swin, we assume that the input size is 224$\times$224 and the smallest windows size is 7. For this reason, there are only two options for the window size at stages 1 to 3, which are 7 and 14, whereas stage 4 should always have 7 as the window size. With that in mind, Table \ref{table:win} shows the latency and accuracy for all possible configurations that we can have from Swin-B with no shifted windows. It is worth mentioning that Case 1, with configuration `7, 7, 7, 7', is the same as Swin-B without shifted windows, which is the same as Case 8 of Table \ref{table:shifton}.

We can first notice from Cases 2 to 4 in Table \ref{table:win} that the most effective stage to use 14 as the window size is stage 3. Increasing the window size to 14 at stage 3 leads to the best latency and accuracy compared to using the window size of 14 at stage 1 or 2.
This would again come from the fact that the stage 3 is the dominant part of Swin-B in terms of  depth.
Using a 14$\times$14 local window at stage 3, we take cross-connection into account among four neighboring 7$\times$7 local windows at stage 2. 
Note that using the larger window size means that we need to handle larger-kernel matrix multiplications, but the number of such matrix multiplications (\textit{i.e.}, the number of non-overlapping windows) gets smaller (refer to Table \ref{table:comp_arch}).
Comparing latency results between Cases 1 and 2, this rather helps reducing the latency. We may claim the same improvement in latency at stage 1 or 2 by using a window size of 14, but considering that those stages are of only depth two, we could not observe meaningful speed-up there. See that Cases 3 and 4 get the same latency as Case 1 up to the first decimal point.

In Cases 5 to 7, we use the 14$\times$14 local window at two stages at the same time. We see that not using the 14$\times$14 local window at stage 3 degrades both accuracy and latency, emphasizing the importance of stage 3 once again. We can also see from Cases 5 and 6 that using the 14$\times$14 local window at stage 1 or 2 in addition to stage 3 meaningfully improves latency over Case 2, resulting in them being the fastest variants.

Looking at Case 8, using a window size of 14 at stages 1 to 3 does not further improve the latency over Case 5 or 6. The accuracy rather slightly decreases. The reason may be that the modeling of cross-window connection is less effective at early stages. From this study, we chose the configuration of \name as Case 5 (as shown in Table \ref{table:comp_arch}), which was one of the best ones in both accuracy and latency.

\begin{table*}[]
\caption{Models trained with ImageNet-1K from scratch. FLOP and parameter counts are measured by \cite{ThanatosShinji}. SwinV2 did not work with this tool so we mark it with `-' here.}\label{table:exp1k}
 \centering
\begin{tabular}{ccccccc}
\hline
\multirow{2}{*}{Case} & \multirow{2}{*}{Model} & \multirow{2}{*}{FLOPs} & \multirow{2}{*}{$\#$ of parameters} & \multirow{2}{*}{Top-1 accuracy (\%)} & \multicolumn{2}{c}{Latency (ms)} \\ \cline{6-7} 
&&&&&{TensorRT (FP16)} & {PyTorch (FP32)} \\ \midrule\midrule[.1em]
1& Swin-B & 15.9G & 88.7M & 83.4 & 2.1 & 14.3 \\ 
2& Swin-B-BR & 15.6G & 88.7M & 83.2 & 1.8 & 15.3 \\ 
3& SwinV2-B & - & - & 83.8 & 3.5 & 21.5 \\ 
4& \name-B & 16.8G & 99.4M & 83.8 & 2.0 & 12.6 \\ \hline

5& \name-T & 5.0G & 31.6M & 82.1 & 0.9 & 6.7 \\ 
6& \name-S & 9.7G & 58.3M & 83.6 & 1.7 & 12.6 \\ \hline

7& \name-T-BR & 4.8G & 31.6M & 82.1 & 0.8 & 7.0 \\ 
8& \name-S-BR & 9.5G & 58.3M & 83.6 & 1.4 & 13.2 \\ 
9& \name-B-BR & 16.4G & 99.4M & 83.7 & 1.7 & 13.2 \\ \hline

10& \name-B-DR10 & 11.3G & 69.3M & 83.5 & 1.4 & 9.3 \\ 
11& \name-B-DR12 & 12.7G & 76.8M & 83.8 & 1.5 & 9.7 \\ 
12& \name-B-DR14 & 14.0G & 84.4M & 83.8 & 1.7 & 10.7 \\ 
13& \name-B-DR16 & 15.4G & 91.9M & 83.8 & 1.9 & 11.6 \\ \hline

14& \name-B-BR-DR12 & 12.4G & 76.9M & 83.3 & 1.3 & 10.1 \\ 
15& \name-B-BR-DR14 & 13.7G & 84.4M & 83.7 & 1.4 & 11.2 \\ 
16& \name-B-BR-DR16 & 15.1G & 91.9M & 83.8 & 1.6 & 12.2 \\ \hline

\end{tabular}
\end{table*}

\subsection{Comparison between Swin and \name}\label{sec:comp_with_swin}

Table \ref{table:exp1k} lists all variants of \name and some of Swin family that we trained from scratch. First, from Cases 1 and 6, we can compare \name with Swin for the Base (B) variant. Although \name-B has more FLOPs and parameters than Swin-B, we can see that \name-B is faster than Swin-B at inference using either PyTorch (12.6 ms vs. 14.3 ms) or TensorRT (2.0ms vs. 2.1ms). From the study in Table \ref{table:win}, we understand this happens because \name-B has a smaller number of non-overlapping windows at stages 2 and 3, although each window is larger in \name-B.

We can also note that \name-B achieves better accuracy than Swin-B. This implies that even without using the shifted windows, changing the size of the local window at certain stages can well model cross-connection among neighboring windows.
Consistently, \name-T and \name-S in Cases 5 and 6 also achieve better accuracy than Swin's corresponding variants (not shown here; Refer to \cite{liu2021swin}).

We also observed that for an input size of 224$\times$224, SwinV2-B \cite{DBLP:journals/corr/abs-2111-09883} gets the same accuracy as \name-B, but its latency is significantly slower. Thus, for latency-critical applications, \name would be a better choice than SwinV2.

\subsection{BatchNorm/ReLU (BR) variants}\label{sec:exp_br}
Replacing LayerNorm and GELU in \name with BatchNorm and ReLU, respectively, we get the variants in Cases 7 to 9 in Table \ref{table:exp1k}. We first notice that the accuracy degradation that occurs with these replacements is trivial. Namely, only \name-B-BR has slightly lower accuracy than \name-B, while others hold the same accuracy as their corresponding models.
In regards to latency, BR variants achieve meaningful speed gain in TensorRT, although not in Pytorch. Nonetheless, considering that TensorRT is a \textit{de facto} standard for deploying a deep learning model, BR variants would be good alternatives in case of latency-critical applications. It is also worth noting from Case 2 that simply applying BR modification to the original Swin-B does not yield similar accuracy or latency as compared to \name-B-BR.

\subsection{Depth reduction (DR$x$) variants}\label{sec:exp_dx}

Cases 10 to 13 in Table \ref{table:exp1k} show the DR$x$ variants of \name-B. Not to mention, D$10$, D$12$, D$14$, and D$16$ variants of \name-B reduce FLOPs and the number of parameters, thereby improving the latency from \name-B. See that in Case 11, \name-B-DR12 has even lower FLOPs than Swin-B and its TensorRT runtime is reduced from 2.0 ms to 1.5 ms when compared to \name-B. In regards to accuracy, we can see that it stays the same as \name-B. This implies that with our size-varying window, we may not need such deep depth of Swin at stage 3.

From Cases 14 to 16, we can also see that the combination of BR and DR$x$ can still result in superior accuracy compared to Swin-B, while improving latency further. For example, \name-B-BR-DR14 has an accuracy of 83.7\% and latency of 1.4 ms, compared to 83.4\% and 2.1 ms from Swin-B.
Note in Cases 1 and 14 that by sacrificing a little bit of accuracy (from 83.4\% to 83.3\%), \name-B-BR-DR12 can achieve significant reduction in latency (from 2.1 ms to 1.3 ms, which is about 38\% reduction from Swin-B). These kinds of \name variants could be attractive alternatives for Swin in  situations where latency is more important than accuracy.

\section{Conclusion}\label{sec:conclusion}

This paper presents \name, which attempts to improve latency over Swin Transformer by reducing memory traffic incurred by shifted window scheme. Instead, \name varies the size of windows over stages, which mimics the mechanism of the shifted windows. This simple technique is shown to offer reduced latency and better accuracy compared to its Swin counterpart. We also show that further speedup can be achieved by using simpler operations and shallower blocks without accuracy loss. 
Therefore, the proposed model is particularly suitable for deployment in production with improved efficiency.

In future work, we plan on applying \name to other vision tasks such as object detection and semantic segmentation with larger input resolution. More optimizations, such as dynamic window size across different stages, will also be investigated to further improve GPU utilization for inference.

{\small
\bibliographystyle{unsrt}
\bibliography{egbib}

\begin{thebibliography}{10}

\bibitem{liu2021swin}
Ze~Liu, Yutong Lin, Yue Cao, Han Hu, Yixuan Wei, Zheng Zhang, Stephen Lin, and
  Baining Guo.
\newblock Swin transformer: Hierarchical vision transformer using shifted
  windows.
\newblock In {\em Proceedings of the IEEE/CVF International Conference on
  Computer Vision (ICCV)}, pages 10012--10022, October 2021.

\bibitem{NIPS2012_c399862d}
Alex Krizhevsky, Ilya Sutskever, and Geoffrey~E Hinton.
\newblock Imagenet classification with deep convolutional neural networks.
\newblock In F.~Pereira, C.J. Burges, L.~Bottou, and K.Q. Weinberger, editors,
  {\em Advances in Neural Information Processing Systems}, volume~25. Curran
  Associates, Inc., 2012.

\bibitem{DBLP:journals/corr/SimonyanZ14a}
Karen Simonyan and Andrew Zisserman.
\newblock Very deep convolutional networks for large-scale image recognition.
\newblock {\em CoRR}, abs/1409.1556, 2014.

\bibitem{DBLP:journals/corr/SzegedyLJSRAEVR14}
Christian Szegedy, Wei Liu, Yangqing Jia, Pierre Sermanet, Scott~E. Reed,
  Dragomir Anguelov, Dumitru Erhan, Vincent Vanhoucke, and Andrew Rabinovich.
\newblock Going deeper with convolutions.
\newblock {\em CoRR}, abs/1409.4842, 2014.

\bibitem{DBLP:journals/corr/HeZRS15}
Kaiming He, Xiangyu Zhang, Shaoqing Ren, and Jian Sun.
\newblock Deep residual learning for image recognition.
\newblock {\em CoRR}, abs/1512.03385, 2015.

\bibitem{DBLP:journals/corr/HanMD15}
Song Han, Huizi Mao, and William~J. Dally.
\newblock Deep compression: Compressing deep neural network with pruning,
  trained quantization and huffman coding.
\newblock In Yoshua Bengio and Yann LeCun, editors, {\em 4th International
  Conference on Learning Representations, {ICLR} 2016, San Juan, Puerto Rico,
  May 2-4, 2016, Conference Track Proceedings}, 2016.

\bibitem{DBLP:journals/corr/HowardZCKWWAA17}
Andrew~G. Howard, Menglong Zhu, Bo~Chen, Dmitry Kalenichenko, Weijun Wang,
  Tobias Weyand, Marco Andreetto, and Hartwig Adam.
\newblock Mobilenets: Efficient convolutional neural networks for mobile vision
  applications.
\newblock {\em CoRR}, abs/1704.04861, 2017.

\bibitem{vit}
Alexey Dosovitskiy, Lucas Beyer, Alexander Kolesnikov, Dirk Weissenborn,
  Xiaohua Zhai, Thomas Unterthiner, Mostafa Dehghani, Matthias Minderer, Georg
  Heigold, Sylvain Gelly, Jakob Uszkoreit, and Neil Houlsby.
\newblock An image is worth 16x16 words: Transformers for image recognition at
  scale.
\newblock In {\em International Conference on Learning Representations}, 2021.

\bibitem{DBLP:journals/corr/abs-2111-09883}
Ze~Liu, Han Hu, Yutong Lin, Zhuliang Yao, Zhenda Xie, Yixuan Wei, Jia Ning, Yue
  Cao, Zheng Zhang, Li~Dong, Furu Wei, and Baining Guo.
\newblock Swin transformer {V2:} scaling up capacity and resolution.
\newblock {\em CoRR}, abs/2111.09883, 2021.

\bibitem{DBLP:journals/corr/abs-1810-04805}
Jacob Devlin, Ming{-}Wei Chang, Kenton Lee, and Kristina Toutanova.
\newblock {BERT:} pre-training of deep bidirectional transformers for language
  understanding.
\newblock {\em CoRR}, abs/1810.04805, 2018.

\bibitem{radford2018improving}
Alec Radford, Karthik Narasimhan, Tim Salimans, and Ilya Sutskever.
\newblock Improving language understanding by generative pre-training.
\newblock 2018.

\bibitem{DBLP:journals/corr/abs-2005-14165}
Tom~B. Brown et~al.
\newblock Language models are few-shot learners.
\newblock {\em CoRR}, abs/2005.14165, 2020.

\bibitem{xie2021segformer}
Enze Xie, Wenhai Wang, Zhiding Yu, Anima Anandkumar, Jose~M. Alvarez, and Ping
  Luo.
\newblock {SegFormer}: Simple and efficient design for semantic segmentation
  with transformers.
\newblock In {\em Advances in Neural Information Processing Systems 34
  pre-proceedings (NeurIPS)}, 2021.

\bibitem{HochSchm97}
Sepp Hochreiter and Jürgen Schmidhuber.
\newblock Long short-term memory.
\newblock {\em Neural Computation}, 9(8):1735--1780, 1997.

\bibitem{chowdhery2022palm}
Aakanksha Chowdhery et~al.
\newblock Palm: Scaling language modeling with pathways, 2022.

\bibitem{tensorrt}
NVIDIA TensorRT.
\newblock https://developer.nvidia.com/tensorrt.

\bibitem{vaswani2017attentionisallyouneed}
Ashish Vaswani, Noam Shazeer, Niki Parmar, Jakob Uszkoreit, Llion Jones,
  Aidan~N Gomez, {\L}ukasz Kaiser, and Illia Polosukhin.
\newblock Attention is all you need.
\newblock In {\em Advances in neural information processing systems}, pages
  5998--6008, 2017.

\bibitem{DBLP:journals/corr/abs-1910-10683}
Colin Raffel, Noam Shazeer, Adam Roberts, Katherine Lee, Sharan Narang, Michael
  Matena, Yanqi Zhou, Wei Li, and Peter~J. Liu.
\newblock Exploring the limits of transfer learning with a unified text-to-text
  transformer.
\newblock {\em CoRR}, abs/1910.10683, 2019.

\bibitem{DBLP:journals/corr/abs-1901-02860}
Zihang Dai, Zhilin Yang, Yiming Yang, Jaime~G. Carbonell, Quoc~V. Le, and
  Ruslan Salakhutdinov.
\newblock Transformer-xl: Attentive language models beyond a fixed-length
  context.
\newblock {\em CoRR}, abs/1901.02860, 2019.

\bibitem{DETR}
Nicolas Carion, Francisco Massa, Gabriel Synnaeve, Nicolas Usunier, Alexander
  Kirillov, and Sergey Zagoruyko.
\newblock End-to-end object detection with transformers.
\newblock {\em CoRR}, abs/2005.12872, 2020.

\bibitem{SETR}
Sixiao Zheng, Jiachen Lu, Hengshuang Zhao, Xiatian Zhu, Zekun Luo, Yabiao Wang,
  Yanwei Fu, Jianfeng Feng, Tao Xiang, Philip H.~S. Torr, and Li~Zhang.
\newblock Rethinking semantic segmentation from a sequence-to-sequence
  perspective with transformers.
\newblock {\em CoRR}, abs/2012.15840, 2020.

\bibitem{bai2019}
Junjie Bai, Fang Lu, Ke~Zhang, et~al.
\newblock {ONNX}: Open neural network exchange.
\newblock \url{https://github.com/onnx/onnx}, 2019.

\bibitem{dest}
John Yang, Le~An, Anurag Dixit, Jinkyu Koo, and Su~Inn Park.
\newblock Depth estimation with simplified transformer, 2022.

\bibitem{deng2009imagenet}
Jia Deng, Wei Dong, Richard Socher, Li-Jia Li, Kai Li, and Li~Fei-Fei.
\newblock Imagenet: A large-scale hierarchical image database.
\newblock In {\em 2009 IEEE conference on computer vision and pattern
  recognition}, pages 248--255, 2009.

\bibitem{ThanatosShinji}
ThanatosShinji.
\newblock {onnx-tool}.
\newblock \url{https://github.com/ThanatosShinji/onnx-tool}, 2023.

\end{thebibliography}
}

\end{document}